\documentclass{article}

\usepackage[preprint, nonatbib]{neurips_2019}
\usepackage{microtype}
\usepackage{booktabs} 
\usepackage{hyperref}
\usepackage{amsmath}
\usepackage{amssymb}
\usepackage{verbatim}
\usepackage{xcolor}
\usepackage{float}
\usepackage{subfig}
\usepackage{xstring}
\usepackage{tikz}
\usetikzlibrary{automata,calc,backgrounds,arrows,positioning, shapes.misc,quotes,arrows.meta}
\usepackage{pgfplots}
\usepackage{mathtools}
\usepackage{array}
\usepackage{multirow}
\usepackage[utf8]{inputenc} 
\usepackage[T1]{fontenc}    
\usepackage{url}            
\usepackage{nicefrac}       

\newcommand{\x}{{\pmb{x}}}

\newcommand{\z}{{\pmb{z}}}
\newcommand{\zet}{{\pmb{\zeta}}}

\newcommand{\logit}{{\pmb{l}}}

\newcommand{\h}{{\pmb{h}}}

\newcommand{\uu}{$\uparrow$}
\newcommand{\dd}{$\downarrow$}

\title{PixelVAE++: Improved PixelVAE with Discrete Prior}

\author{%
  Hossein Sadeghi \\
  D-Wave Systems Inc.\\
  \texttt{hsadeghi@dwavesys.com} \\
  \And
  Evgeny Andriyash \\
  D-Wave Systems Inc. \\
  \texttt{eandriyash@dwavesys.com} \\
  \And
  Walter Vinci \\
   D-Wave Systems Inc.\\
  \And
  Lorenzo Buffoni \\
   D-Wave Systems Inc.\\
  University of Florence \\
  \texttt{lorenzo.buffoni@unifi.it} \\
  \And
  Mohammad H. Amin \\
  D-Wave Systems Inc. \\
  Simon Fraser University \\
  \texttt{amin@dwavesys.com} \\
}

\begin{document}

\maketitle

\begin{abstract}
  Constructing powerful generative models for natural images is a challenging task. PixelCNN models capture details and local information in images very well but have limited receptive field. Variational autoencoders with a factorial decoder can capture global information easily, but they often fail to reconstruct details faithfully. PixelVAE combines the best features of the two models and constructs a generative model that is able to learn local and global structures. Here we introduce PixelVAE++, a VAE with three types of latent variables and a PixelCNN++ for the decoder. We introduce a novel architecture that reuses a part of the decoder as an encoder. We achieve the state of the art performance on binary data sets such as MNIST and Omniglot and achieve the state of the art performance on CIFAR-10 among latent variable models while keeping the latent variables informative.
\end{abstract}

\section{Introduction}

Generative modeling of images \cite{van2016conditional, goodfellow_generative_2014,  gregor2016towards, kingma2016improved, dinh2016density}, audio \cite{oord2016wavenet, mehri2016samplernn}, and videos \cite{kalchbrenner2017video, finn2016unsupervised} has advanced remarkably in the past few years and have resulted in great applications \cite{ledig2017photo, isola2017image}. Many machine learning tasks such as domain adaptation \cite{hoffman2013efficient, ghifary2014domain, bousmalis2017unsupervised, wang2018deep} and few-shot learning \cite{ravi2016optimization, santoro2016one, snell2017prototypical} rely on learned representations of observations.
The goal of generative learning is to model the distribution of unseen observations. This task is achieved with some degree of success by latent variable models such as variational autoencoder (VAE) \cite{kingma2013auto, rezende2014stochastic, salimans2015markov, kingma2016improved}, flow-based models \cite{kingma2018glow, rezende2015variational, dinh2014nice, dinh2016density}, energy-based model \cite{du2019implicit}, autoregressive models such as PixelCNN \cite{salimans2017pixelcnn++, van2016conditional, van2016pixel, oord2016wavenet, chen2017pixelsnail, germain2015made}, and lately, by the combination of latent-variable and autoregressive models, PixelVAE \cite{gulrajani2016pixelvae, chen2016variational}.

While PixelCNN variants are able to model local correlations successfully, large models and self-attention mechanisms are necessary to capture long-range correlations \cite{chen2017pixelsnail}. VAEs, on the other hand, are able to capture global information in the latent space but generally fail to model local structure variations. Modelling the local structure, while not important for downstream tasks such as classification \cite{bengio2013deep}, is essential for building a powerful generative model. Early models with autoregressive decoder found that the autoregressive part learns all the features of the data and the latent variables are not used \cite{fabius2014variational, bowman2015generating, chung2014empirical, serban2016building, serban2017hierarchical, fraccaro2016sequential}. Using a shallow PixelCNN decoder, PixelVAE \cite{gulrajani2016pixelvae} was able to achieve a generative performance comparable with PixelCNN, with hierarchical latent variables that meaningfully control the global features of generated images. However, in \cite{chen2016variational}, authors argued that in principle the optimal point of a PixelVAE is where all information is modelled by the autoregressive part, but in practice this may not be achievable.

In this manuscript, we present PixelVAE++, a model that improves performance of PixelCNN++ and achieves state of the art performance on a few data sets.
Our contribution can be summarized as follows:
\begin{itemize}
    \item We combine PixelCNN++ architecture with VAEs. We show that using latent variables improves performance.
    \item We implement a hierarchical encoder that shares most of its parameters with the autoregressive decoder.
    \item We show that the choice of prior is essential in building a useful latent variable model. In particular, we use discrete latent variables with an RBM prior.
\end{itemize}

\section{Background}
PixelCNN \cite{van2016conditional} is an autoregressive generative model with a tractable likelihood. The model fully factorizes the probability density function as follow: 
\begin{equation}
    p(\x) = \prod_i p(x_i|\x_{<i}),
\end{equation} 
where $\x_{<i}$ is the set of all $x_j$ with $j<i$.
The conditional distributions $p(x_i|\x_{<i})$ are parameterized by convolutional neural networks. The functional forms of these conditionals are very flexible, making PixelCNN a powerful generative model. We used the most recent implementation of PixelCNN++ \cite{salimans2017pixelcnn++} that removes the blind spots and achieves better performance than previous PixelCNN models.

PixelVAE \cite{gulrajani2016pixelvae} is a VAE with PixelCNN as its decoder. Since its log-likelihood is intractable for large number of latent variables, the evidence lower bound (ELBO) is minimized as the objective function \cite{kingma2013auto}. 
\begin{equation}
     \mathcal{L}(\x) = \mathbb{E}_{\z \sim q(\z|\x)}\left[\log p_{\theta}(\x|\z) - \log \frac{p_{\theta}(\z)}{q_{\phi}(\z|\x)}\right] \leq \log p(\x)
    \label{elbo}
\end{equation}
where $p_\theta(\z)$ is the prior distribution over the latent variables $\z$, which together with conditional probability $p_\theta(\x|\z)$ forms the generative model $p_\theta(\x, \z)$ with parameters $\theta$, and $q_\phi(\z|\x)$ is the approximating posterior represented as a neural network with parameters $\phi$. 

The conditional probability (decoder) and the approximating posterior (encoder) are trained end-to-end to optimize the generative and inference models jointly. Many researches around VAE models deal with building more expressive distributions for all its components \cite{kingma2016improved, tomczak2017vae, vahdat2018dvae++, gulrajani2016pixelvae, chen2016variational}. PixelVAE in particular adds more power to the structure of the decoder by replacing conditionally independent dimensions with a fully autoregressive structure.

\subsection{Discrete Variational Autoencoder (DVAE)}
\label{DVAE}

Restricted Boltzmann machines model multi-modal distributions and can capture complex relationship in data \cite{le2008representational}. Using RBM as a prior in a generative model has been pioneered by \cite{rolfe2016discrete} and later improved in \cite{vahdat2018dvae++, vahdat2018dvae, vahdat2019learning, khoshaman2018gumbolt}. In order to back-propagate through the approximating posterior of discrete binary variables, we employ a continuous relaxation estimator that trades bias for variance \cite{bengio2013estimating, raiko2014techniques}, in particular the Gumbel-Softmax estimator \cite{jang2016categorical, maddison2016concrete}. Reference \cite{khoshaman2018gumbolt} showed that a relaxed (biased) objective can be used to train a DVAE with an RBM prior. The relaxed objective can be expressed as 
\begin{equation}\label{eq:relaxation}
    \mathcal{L}(\x) = \mathbb{E}_{\rho \sim \mathbb{U}}\left[\log p_{\theta}(\x|\zet_{\phi,\rho}) - \log \frac{p_{\theta}(\zet_{\phi,\rho})}{q_{\phi}(\zet_{\phi,\rho}|\x)}\right]
\end{equation}
where $\rho$ is a random variable drawn from a uniform distribution $\mathbb{U}$, and $\zet$ is the continuous relaxation of the discrete binary latent variables, 
\begin{equation}
    \zet = \sigma\left[\frac{\logit_\phi + \sigma^{-1}(\rho)}{\tau}\right]
\end{equation}
with $\sigma(x) = 1/(1 + e^{-x})$ being the sigmoid function, $\logit_\phi$ is the logit (output of a neural network), and $\tau$ is the temperature that controls the sharpness of the function $\zet$. The continuous variable $\zet$ is equal to the discrete variable $\z$ in the limit $\tau \rightarrow 0$. At the test time only discrete variables are used which can be obtained by setting the temperature to zero.

\subsection{PixelVAE++ Architecture}
For the decoder of the PixelVAE we implement a model similar to PixelCNN++ \cite{salimans2017pixelcnn++} using downward and rightward shifted image and feature maps. Similarly, we use 6 blocks of $n$ ResNet layers, where $n=5$ for 32x32x3 inputs and smaller ($n$=3) for 28x28 inputs. Down sampling is performed between the 1st/2nd and 2nd/3rd blocks using convolutions with strides of 2, and up-sampling is performed between the 4th/5th and 5th/6th blocks. Similar to PixelCNN++, convolutional connections are introduced between early and late layers to ensure that details and fine structures are preserved at construction time. 

For the encoder, we use three groups of latent variables with a factorial Bernoulli distribution. The encoding distribution of the first group, \textit{concatenated latent variables} $\z_1$, is parameterized by convolutional neural networks (CNN) followed by a dense layer. Using deconvolutions with up-sampling, the decoder transforms the stochastic variables to the size of the input and then concatenates them with the input, similar to the original PixelVAE \cite{gulrajani2016pixelvae}.

The parameters of the second encoding distribution is obtained by a separate set of CNN and dense layers. In the decoder, these \textit{conditioning latent variables} $\z_2$ are added to activations before nonlinearities, similar to the conditional PixelCNN \cite{van2016conditional}.

\begin{figure}[h]
 \centering
  \includegraphics[width=0.9\linewidth]{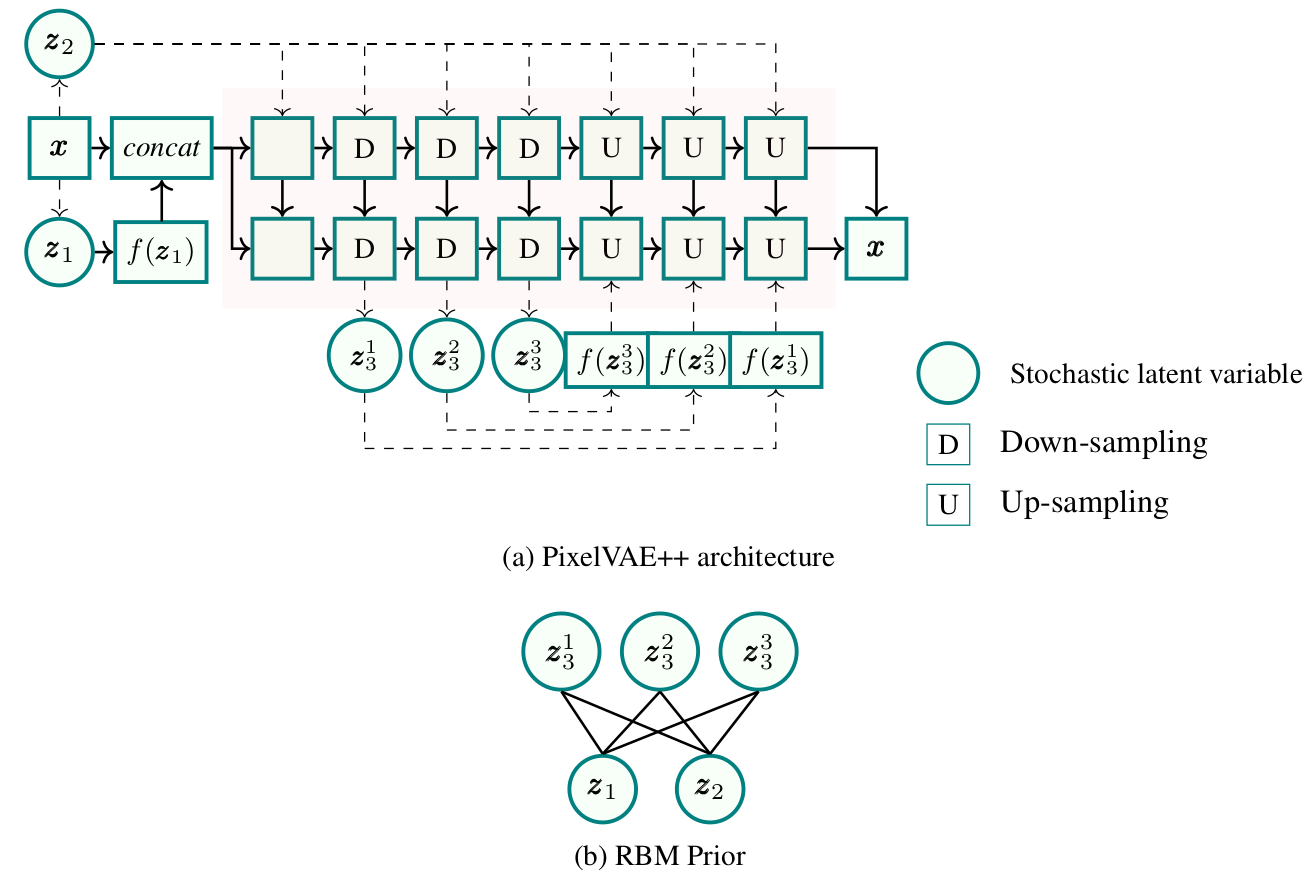}
\caption{The hierarchical structure of PixelVAE++ with RBM prior. Arrows indicate convolutional connection with autoregressive structure. Dashed arrows indicate convolutional connection. The decoder and encoder structures are fused together. The series of down-sampling stages that constitute a of the decoder for $\z_1$ and $\z_2$ variables, acts as an encoder for $\z_3$ variables. Note that the skip connections in the decoder part is omitted in the model graph to reduce clutter.}
  \label{generativeform}
\end{figure}

The architecture of the third group, the {\it shared latent variables} $\z_3$, leverages the autoencoding structure of the PixelCNN++. There are $(n+1) {\times} 3$ layers up to the last down-sampling stage, where n ($n=5$ for CIFAR-10) is the number of ResNet layers per block. Each layer $h_{b, i}$ ($b$-block index, $i$-ResNet layer index) with the size $[B, H, W, N]$, ([batch size, height, width, filters]) is reduced to a $[B, 8, 8, 1]$ by convolution and is followed by a dense layer to 64 variables to produce the logits of the Bernoulli distribution. Each latent variable is then transformed by a dense layer, followed by up-sampling to the same size as $h_{b, i}$. The transformed stochastic variable is concatenated with $h_{b, i}$ and is passed to the corresponding up-sampling block $h_{7-b, n+1-i}$. A gated ResNet combines this layer and the transformed latent variables. A schematic drawing is presented in Fig.~\ref{generativeform}.

\section{Experiments}
We study the performance of PixelVAE++ for density estimation on 2D images. The experiments are performed on statically and dynamically~\cite{lecun1998gradient} binarized MNIST, Omniglot, and Caltech-101 silhouettes~\cite{marlin2010inductive} and CIFAR-10 \cite{krizhevsky2009learning}. For all data sets, we use the standard allocation of training, validation, and test sets. We trained the models for 500 epochs on one GPU for binary data sets and 1000 epochs on 8 GPUS for CIFAR-10. Our goal is to determine whether we can improve the performance of PixelCNN++ with VAE and to understand what effect the use of discrete binary latent variables has on the performance of PixelVAE++, if any.

\subsection{Performance of PixelVAE++}
Some inputs have $28{\times}28$ binary pixel images. We use the same architecture for the MNIST and Omniglot data sets with 3 ResNet layers per block, each having 64 feature maps. Because the Caltech-101 silhouettes data set is smaller, our model with PixelCNN++ decoder using strides of 2 for down-sampling and up-sampling, easily over-fits. We remove the strides and reduce the number of feature maps to 32, but keep the number of ResNet layers the same. We found that for input sizes of 28x28, only three ResNet layers per block was needed to achieve optimal performance. The Log-Likelihood (LL) is evaluated using 1000 importance weighted samples \cite{burda2015importance}. We were able to achieve better performance than what is already published on dynamically binarized MNIST using PixelCNN++ and we further improve the LL using PixelVAE++. The LL reaches the values of -78.00 and -88.29 for MNIST (Dynamic) and Omniglot, respectively. We obtained this results by including only the first group of latent variables, but we also experimented with including all three categories of latent variables. Only the first category, the concatenated variables, was necessary and sufficient for optimal LL in these data sets and adding the other groups did not improve results. We have repeated the experiments for each data set three times and have reported the average. The standard deviation for the mean is $\pm 0.04$, therefore the last digit is uncertain. However, we have followed the norm to report the results for the binary data sets up to the second digit. 

For the binary data sets, we use 400 binary variables in the latent space  all of which belong to concatenated latent variables. The prior consists of an RBM with 200 variables on each side of the bipartite graph. We sample from the RBM during training to computing the gradients of the log-partition function. We use 5000 samples obtained by annealed importance sampling (AIS) \cite{neal2001annealed} with 1000 temperatures steps and 50 MCMC update per step. For the evaluation of the log-partition function, we increase the number of samples and temperature steps tenfold. For training we used batch sizes of 128 samples.

Experiments on CIFAR-10 are done using the same set of hyper-parameters and network architecture as in PixelCNN++ \cite{salimans2017pixelcnn++}, except for the number of filters which we reduce from 160 to 128. We use batch sizes of 8 samples per GPU or 64 samples in total. We only add and tune VAE related parameters and hyper-parameters. Our model has 40\,M parameters, including  6\,M for VAE,  as compared to 54\,M in the original PixelCNN++. Despite the reduced number of parameters, the negative log-likelihood in bpd reaches 2.90. The decoder alone with 128 filters only reaches 2.95 bpd, higher than 2.92 bpd with 160 filters. Increasing the number of filters back to 160 for the VAE model did not result in a better performance than 2.90 bpd. The details of the implementation is outlined in the appendix \ref{app:architecture} for $32{\times}32{\times}3$ inputs. The numbers are reported after averaging three independent runs with a standard deviation of $\pm 0.001$. The standard deviation of log-likelihood for different evaluations of importance weighted sum is negligible.

\begin{table}
\footnotesize
\centering
\begin{tabular}{l r r r r}
\hline
MNIST                & Static          &      & Dynamic         &     \\
                     & LL              & KL   & LL              & KL  \\
\hline   
PixelVAE++ Gaussian  & -78.66          & 6.86 & -78.01          & 4.2 \\
PixelVAE++ RBM       & \textbf{-78.65} & 7.62 & \textbf{-78.00} & 5.05\\
\hline 
VLAE                 & -79.03          &      & -78.53          &     \\
\hline 

\\
                     & OMNIGLOT        &      & Caltech 101      &      \\
                     & LL              & KL   & LL               & KL   \\
\hline   
PixelVAE++ Gaussian  & -88.65          & 1.63 & -79.52           & 4.00 \\
PixelVAE++ RBM       & \textbf{-88.29} & 2.56 & -77.46           & 6.85 \\
\hline 
VLAE                 & -89.83          &      &  \textbf{-77.36} &      \\
\hline

\\
bpd                 & CIFAR10         &      &       &      \\
                    & LL              & KL   &       &      \\
\hline   
PixelVAE++ Gaussian & 2.92            & 0.005&       &      \\
PixelVAE++ RBM      & 2.90            & 0.016&       &      \\
\hline 
VLAE                & 2.95            &      &       &      \\
PixelCNN++          & 2.92            &      &       &      \\
PixelSNAIL          & \textbf{2.85}   &      &       &      \\
\hline 
\\
\end{tabular}
\caption{This table summarizes the results for PixelVAE++ experiments on multiple data sets.}
\label{table:results}
\end{table}

For the CIFAR-10 data set, we use an architecture that includes all three groups of latent variables: 512 concatenated, 128 conditional, and $8{\times}8{\times}3{\times}(n{+}1)$ shared. To construct the RBM, we place the first two groups on one side and the shared variables on the other side of the fully visible RBM. Due to the increased size of the RBM and to reduce the computational cost of AIS we use only 500 samples for training and 5000 samples for evaluation.

\subsection{Conditional image generation}
\begin{figure}[h]
    \centering
    \includegraphics[width=6cm]{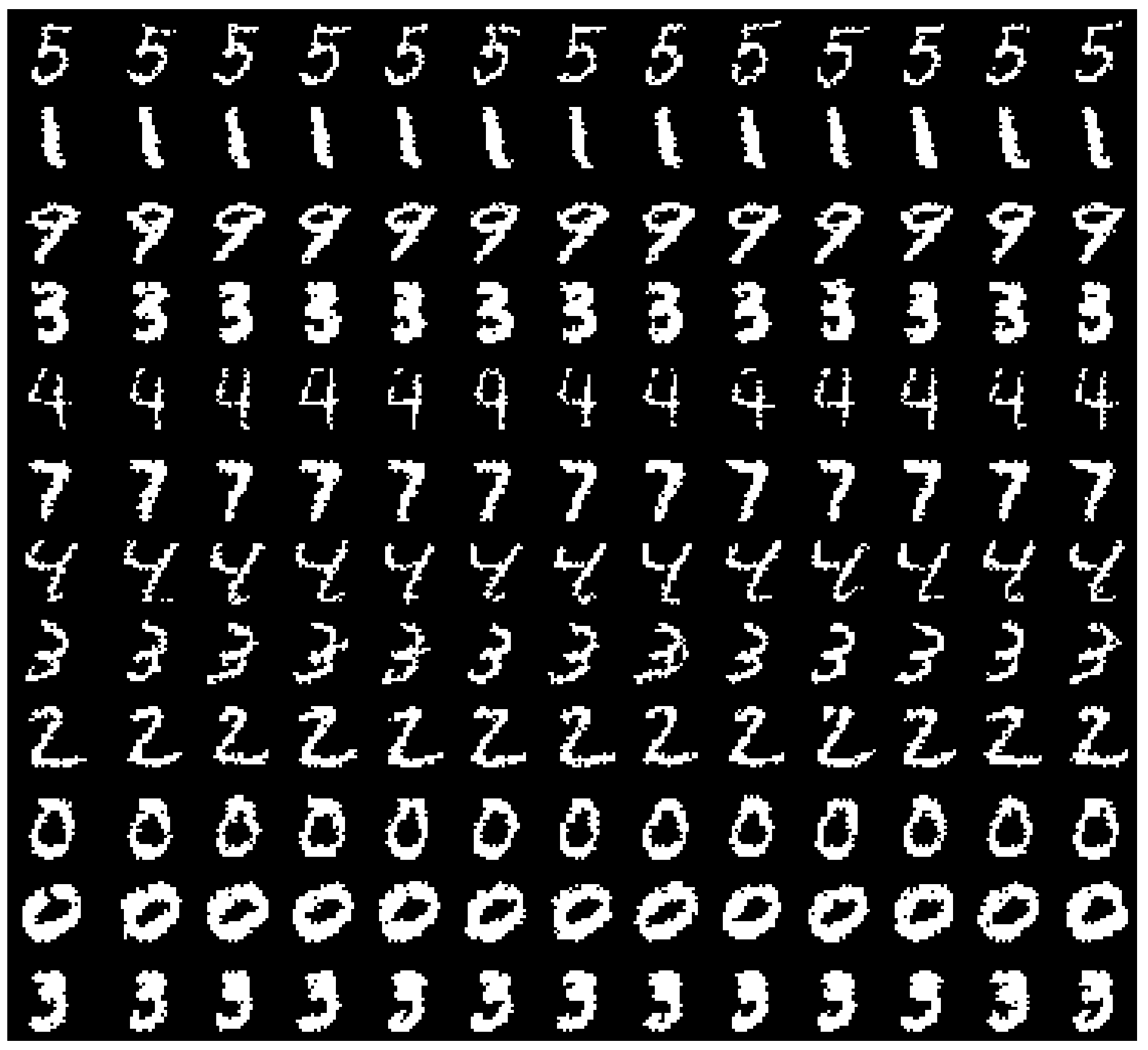}
    \includegraphics[width=6cm]{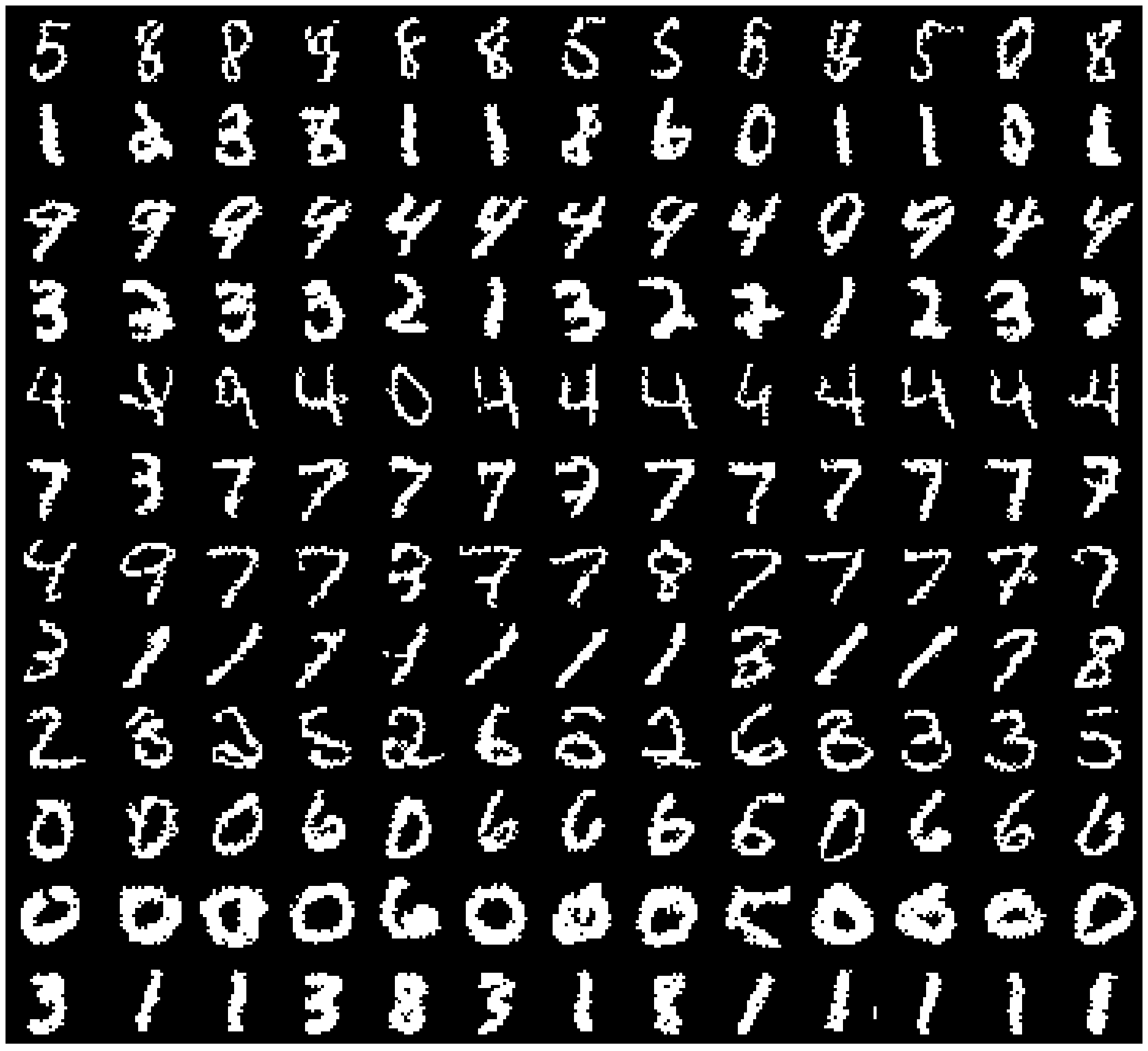}
    \caption{Reconstructed samples for a PixelVAE++ trained on MNIST. Each row is generated from latent variables inferred from the image in the first column. The prior distributions are RBM (left) and Gaussian (right).}
    \label{fig:mnist_cond}
\end{figure}

\begin{figure}[h]
    \centering
    \includegraphics[width=6cm]{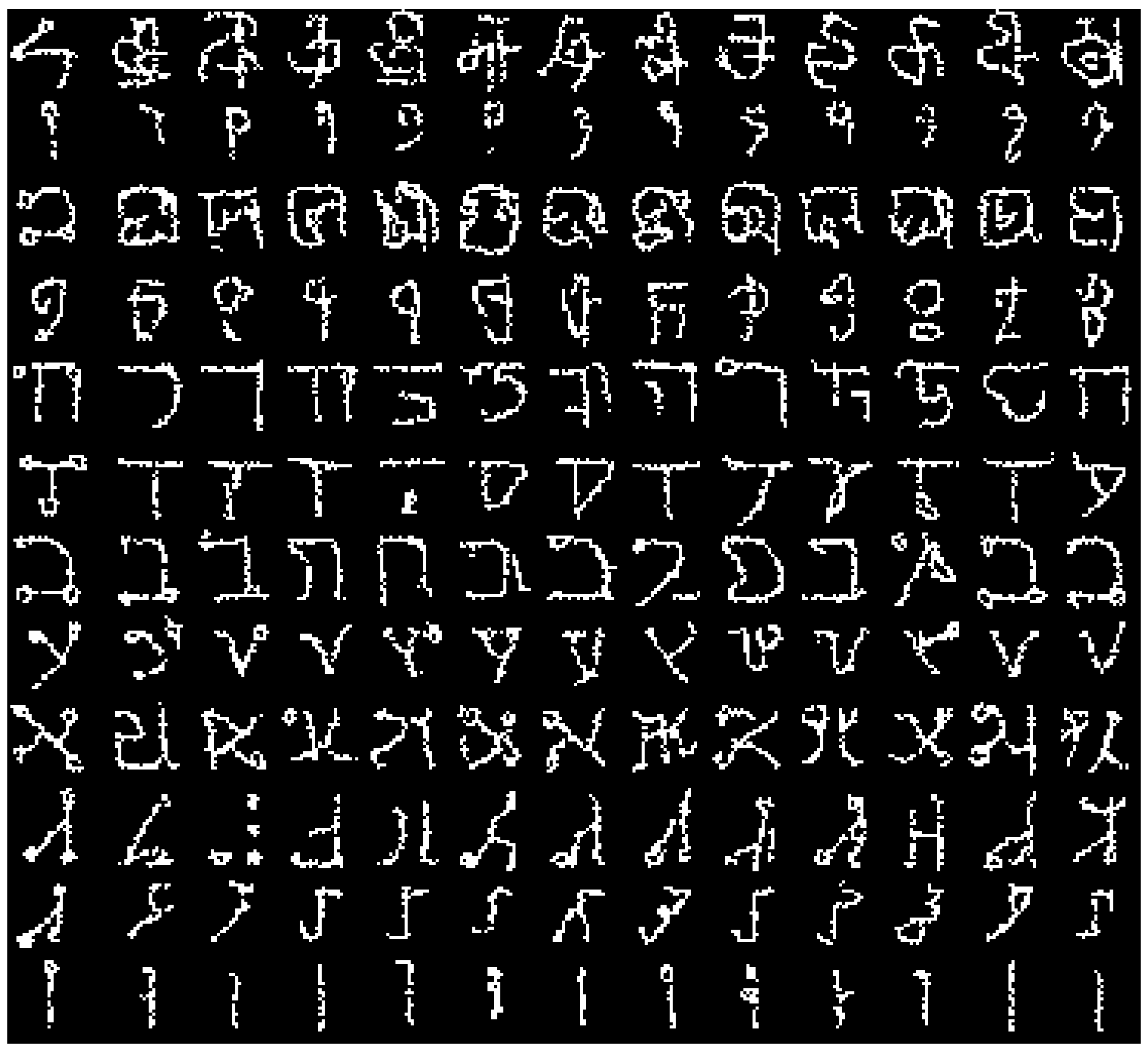}
    \includegraphics[width=6cm]{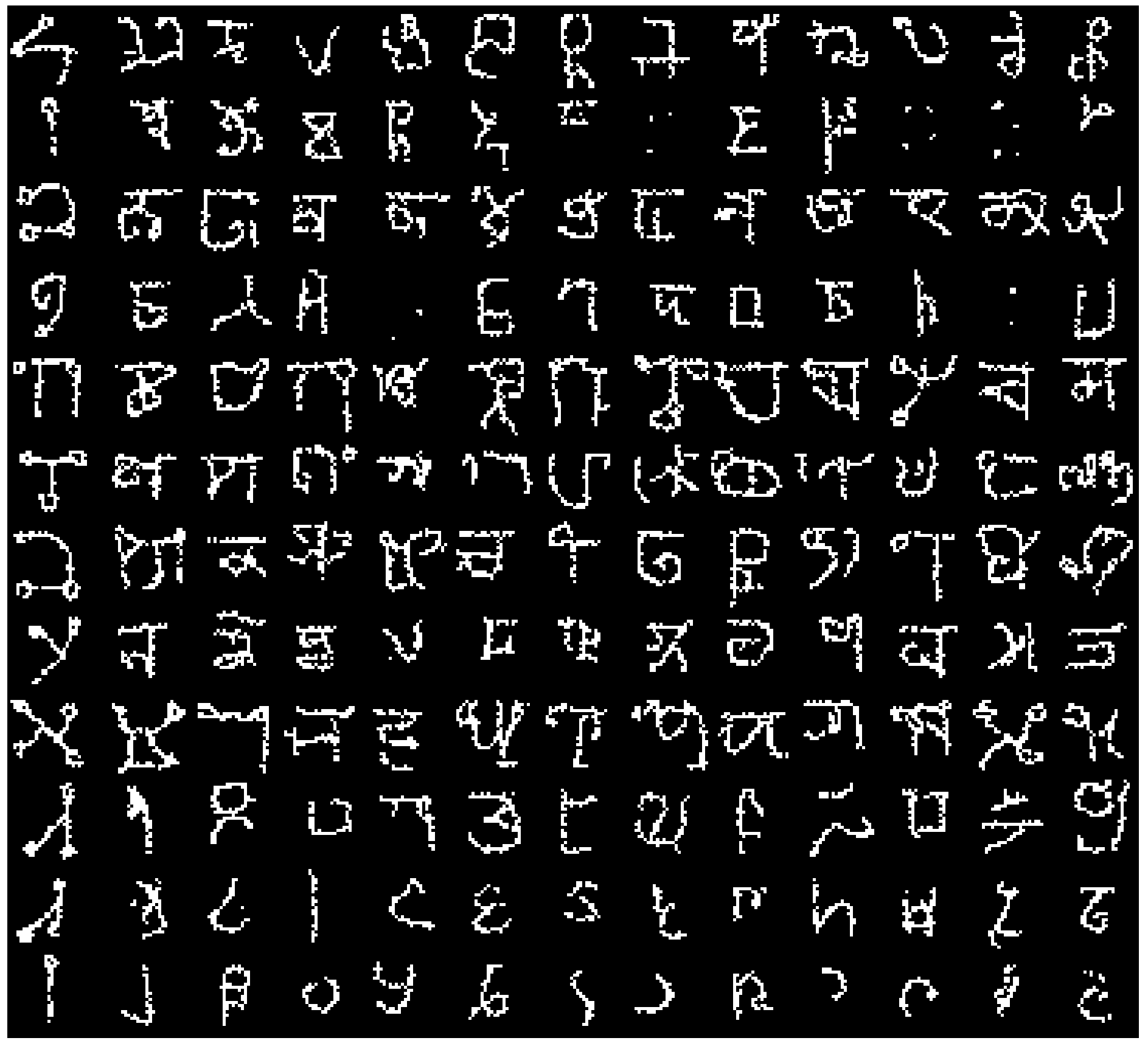}
    \caption{Reconstructed samples for a PixelVAE++ trained on Omniglot. Each row is generated from latent variables inferred from the image in the first column. The prior distributions are RBM (left) and Gaussian (right).}
    \label{fig:omni_cond}

\end{figure}

We perform experiments with PixelVAE++ to assess generation of images  conditional on the latent variables given by the approximating posterior. We generate samples using the latent variables inferred from test images. Figures \ref{fig:mnist_cond} and \ref{fig:omni_cond} show reconstructed images from MNIST and Omniglot test sets for  PixelVAE++ trained with RBM and Gaussian priors. As evident from Fig.~\ref{fig:mnist_cond}, both models with continuous and discrete priors construct the digits very well. Despite having similar LL, the model with the RBM prior has a sharp conditional distribution that captures small variations in digit shape, while the one with the Gaussian prior has much broader conditional, that can cover multiple digit classes. For Omniglot data set (Fig.~\ref{fig:omni_cond}), the model with RBM prior similarly has a sharper conditional, although in this case both RBM and Gaussian prior models span multiple object classes.

The similarity of the reconstructed image and the original one can be measured equivalently by conditional probability, KL divergence of approximating posterior from prior (for a fixed ELBO), and the mutual information between $\x$ and $\z$, all of which change monotonically with respect to each other \cite{alemi2017fixing}. We report the KL divergence in table \ref{table:results}. It is evident from the images and the KL values that PixelVAE++ with RBM prior is able to learn models with sharper decoder distributions and more informative latent space than the one with Gaussian prior.

\begin{figure}[htbp]
    \centering
    \includegraphics[width=6cm,trim={0 0 0 0cm},clip]{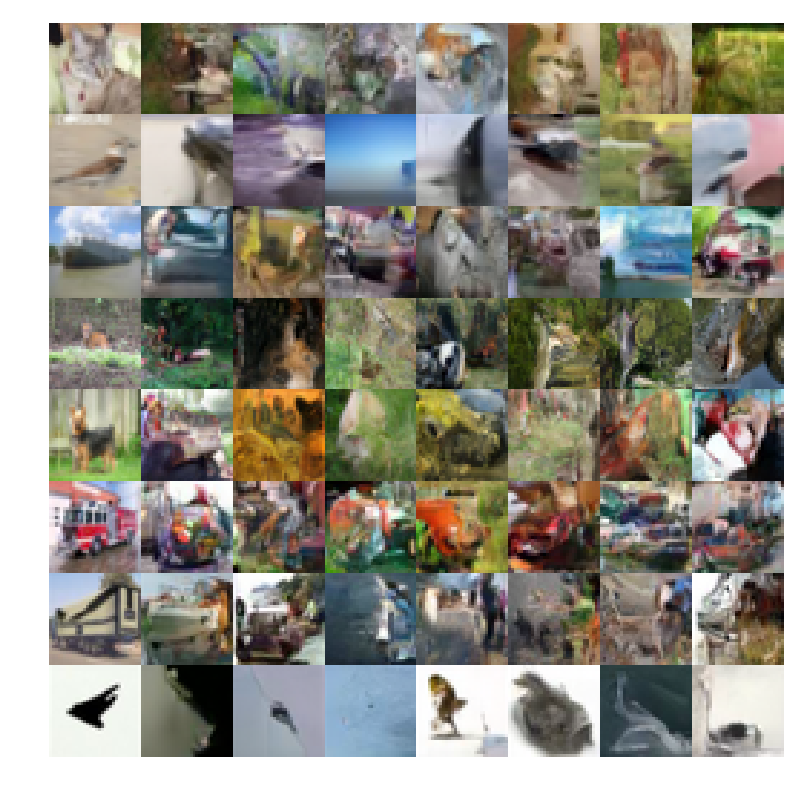}
    \includegraphics[width=6cm,trim={0 0 0 0cm},clip]{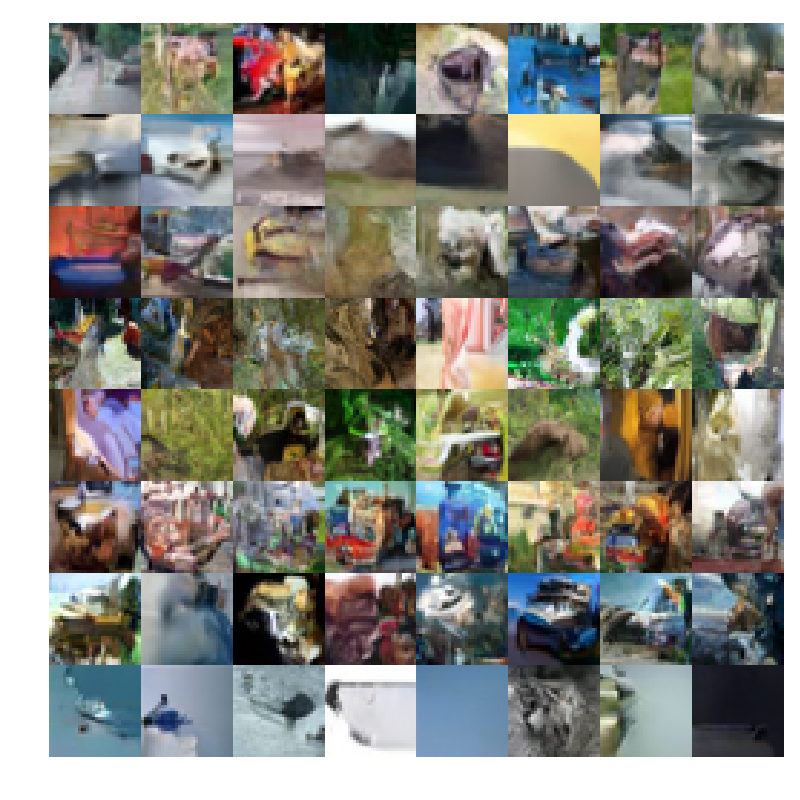}
    \caption{Reconstructed samples for a PixelVAE++ trained on CIFAR-10. Each row is generated from latent variables inferred from the images in the first column. The prior distribution used for training is an RBM. We have sorted rows based on energy distance from the test images and put the first 8 with the closest distance on the left panel and the last 8 on the right panel.}
    \label{fig:cond_cifar}
\end{figure}

For the case of CIFAR-10 images, the decoder distribution becomes rather broad as illustrated in figure \ref{fig:cond_cifar}. To visually highlight how this distribution varies over the data set we choose 128 data points, sample  8 images conditioned on the latent representation of each data point, and then display the ones with the smallest mutual energy distance \cite{salimans2017pixelcnn++} on the left panel, and largest on the right. It is hardly possible to categorize images conditioned on the same latent variable as belonging to the same class, but in some cases it is possible to argue that images are similar in some global appearances such as color and composition.

\section{Conclusions}
We have presented a VAE model using an autoregressive decoder that performs well in generative tasks, in terms of LL, and makes use of the latent variables. This model achieves the best performance among other latent variable models on CIFAR-10 data set by reaching LL of 2.90 bpd using 25\% less parameters than PixelCNN++. 

For the binary data sets, the discrete latent variables capture global features of images (like digit class) while the decoder distribution models local variations.

For more complex natural images, the latent variables help autoregressive model to represent some global features, like color and composition, and achieve better performance in terms of log-likelihood.

\bibliography{discretepixelval}

\newpage
\appendix

\section{Network architecture details}
\label{app:architecture}
The table \ref{table:architecture} outlines the details of the implementation for the CIFAR-10 experiment.

\begin{table}[h]
\centering
\begin{tabular}{ l|c|c}
Module & Type & CIFAR-10 \\
\hline
$q(\z_1|\x)$ & convolutional + dense &  64\dd, 64, 64\dd, 64, 64\dd, 64, 64\dd, 64, 512 FC\\
$f(\z_1)$ & dense + deconvolutional & 512 FC, 64\uu, 64, 64\uu, 64, 64\uu, 64, 16\uu, 16, 3\uu \\
$q(\z_2|\x)$ & convolutional + dense & 1\dd, 128 FC\\
$q(\z_3^1|\h_3^1(\x))$ & convolutional + dense & 1\dd (Kernel = Stride = 4), 64 FC\\
$q(\z_3^2|\h_3^2(\x))$ & convolutional + dense & 1\dd, 64 FC\\
$q(\z_3^3|\h_3^3(\x))$ & convolutional + dense & 1, 64 FC\\
$f(\z_3^1)$ & dense + deconvolutional & 64 FC, 128\uu (Kernel = Stride = 4)\\
$f(\z_3^2)$ & dense + deconvolutional & 64 FC, 128\uu\\
$f(\z_3^3)$ & dense + deconvolutional & 64 FC, 128\\
\end{tabular}
\vspace{0.5cm}
\caption{The details of the implementation. The functions with $q$ are encoding distributions parameterized by neural networks while functions with $f$ are deterministic transformation of latent variables. The arrow signs pointing up and down indicate up-sampling and down-sampling stages in the network. The symbols $h_i^j$ denote the deterministic hidden variables for the latent variables of group i and subgroup j.}
\label{table:architecture}
\end{table}

\section{Improving optimization}
Techniques such as batch normalization~\cite{ioffe2015batch}, dropout \cite{srivastava2014dropout}, weight normalization \cite{salimans2016weight}, and learning-rate decay can significantly improve the performance of a PixelVAE. We evaluate PixelVAE++ by comparing the training of the model to the original PixelCNN++ , both of which include these improvements. For a fair comparison, we apply only those techniques that were also used in \cite{salimans2017pixelcnn++}. We examine whether adding a VAE structure to the PixelCNN++ would improve the performance. 

For VAE specifically, we use optimization methods such as KL annealing~\cite{sonderby2016ladder} which is shown to prevent the approximating posterior from falling into a local minimum.

When we use RBM as the prior, the KL term of the VAE doesn't have a closed form. We reduce the variance of the optimization by neglecting the derivatives of the form $\partial_\phi q_\phi(\z|x)$ and instead the encoder is only trained by the path derivatives \cite{roeder2017sticking}.

\section{Temperature annealing}
The temperature $\tau$ for continuous relaxation (\ref{eq:relaxation}) determines the trade-off between bias and variance. While low temperature relaxation is less biased, its derivative has high variance. We find that a temperature around 0.25 is sufficient to achieve best performance for all experiments. If the temperature is too high (0.5), the mismatch between the smoothed posterior and the RBM grows rapidly. If the temperature is too low (0.1) the high variance of the gradient prevents convergence to below 2.98 bpd in CIFAR-10 experiment.

We experimented with annealing the temperature from high to low and vice versa. The intuition for annealing to high temperature is to reduce variance towards the end of the optimization, and lowering temperature aims at reducing the bias towards the end while keeping the beginning of the training high variance in order to help keeping away from a local minima. In practice however, we find decreasing temperature to degrade performance and increasing it to have no effect compared to training with the final temperature all the way through. This is perhaps due to the high variance of training VAEs and therefore adding the variance due to low temperature has no meaningful effect.

\section{Image generation}
\label{app:conditional}
We evaluate the generated samples conditional on the prior distribution. To sample from the RBM prior, we perform 50000 Gibbs block sampling updates. While the cost of these updates for 1792 variables may be too much for the training phase, it is negligible compared to the autoregressive generation with deep neural networks of 40M variables. With a fixed RBM samples, we then generate multiple samples from the decoder. 

\begin{figure}[h]
    \centering
    \includegraphics[width=6cm]{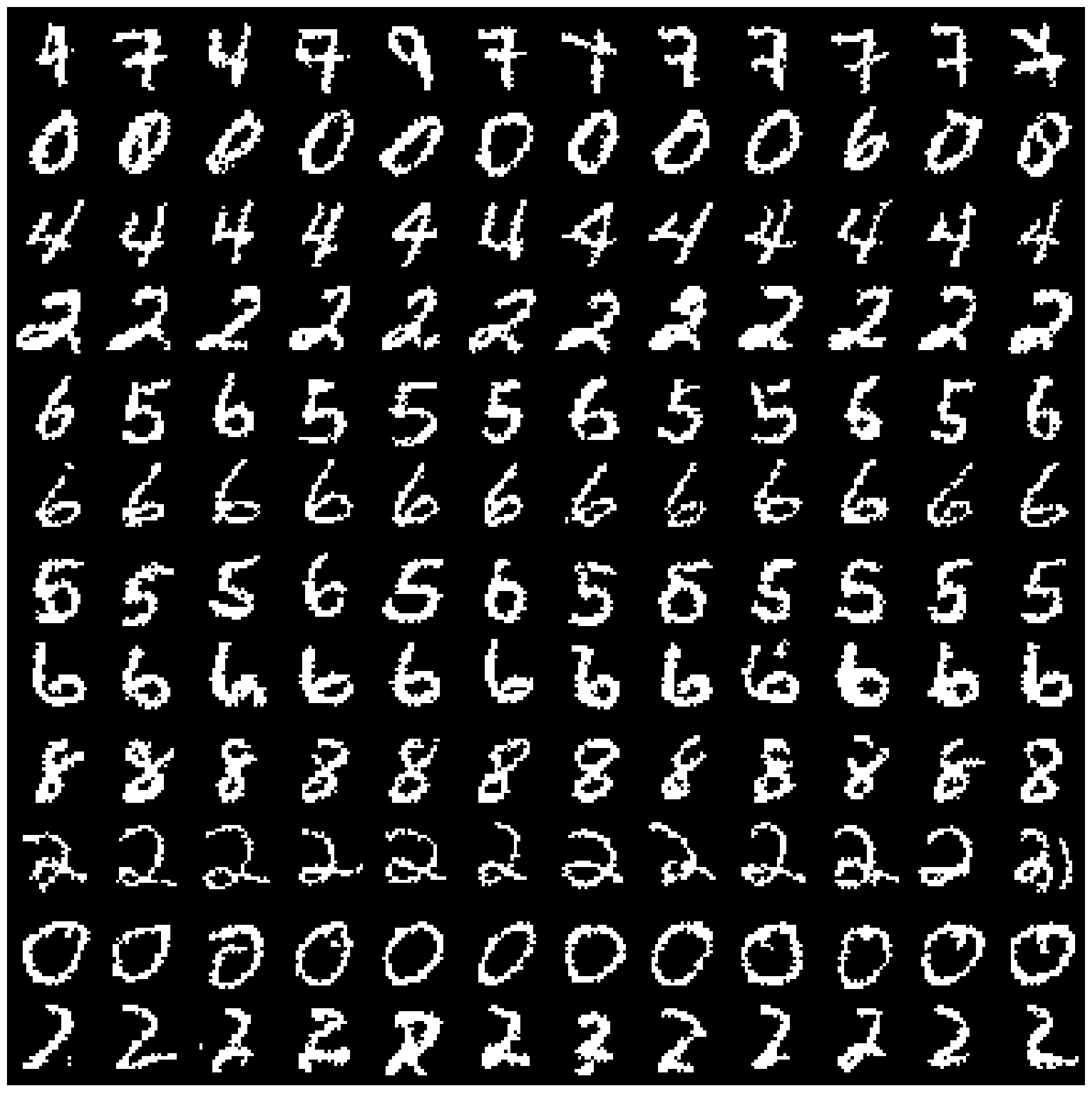}
    \includegraphics[width=6cm]{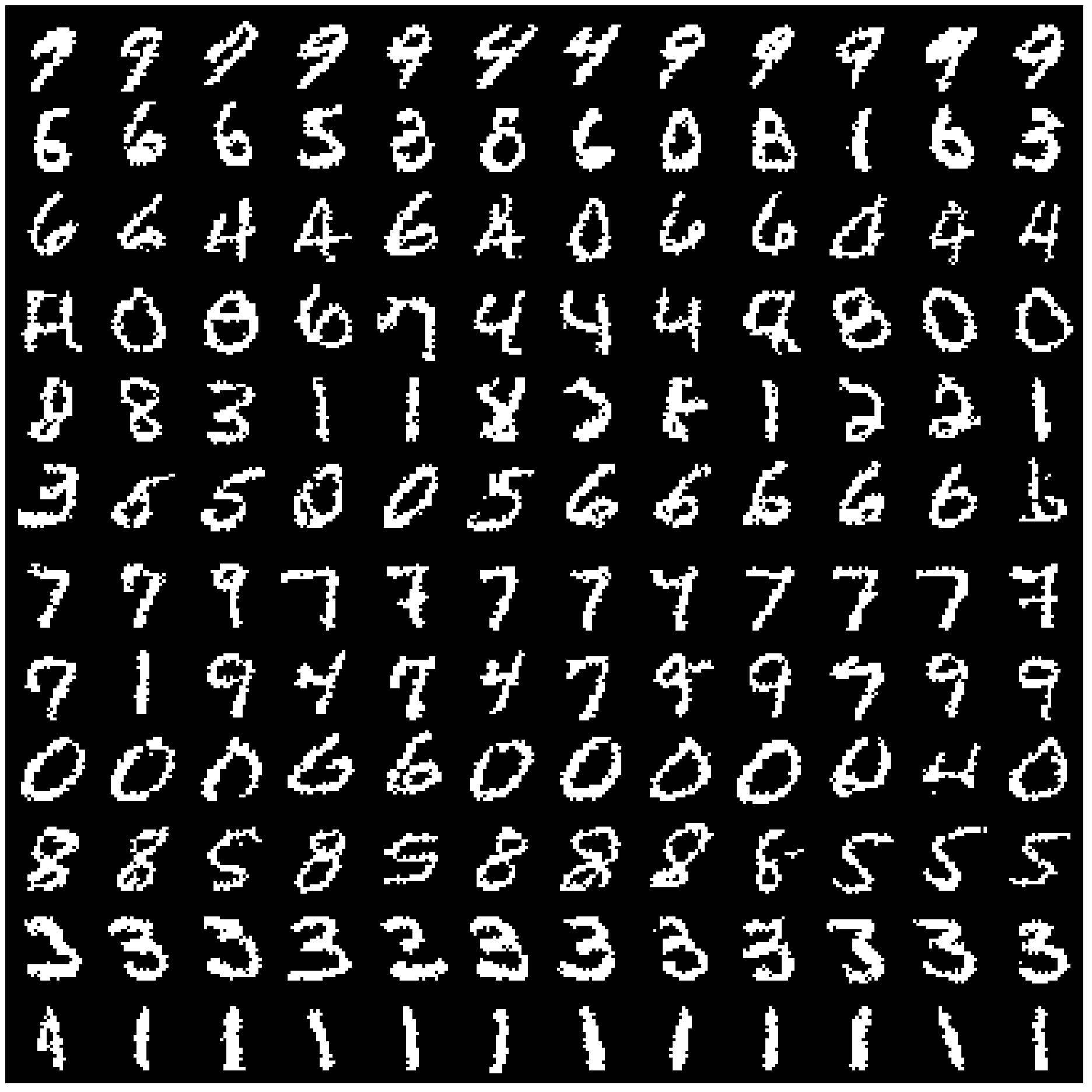}
    \caption{Generated samples for a PixelVAE++ trained on MNIST. Each row is generated from latent variables sampled from the prior distribution. The prior distributions are RBM (left) and Gaussian (right).}
    \label{fig:mnist_gen}
\end{figure}

\begin{figure}[h]
    \centering
    \includegraphics[width=6cm]{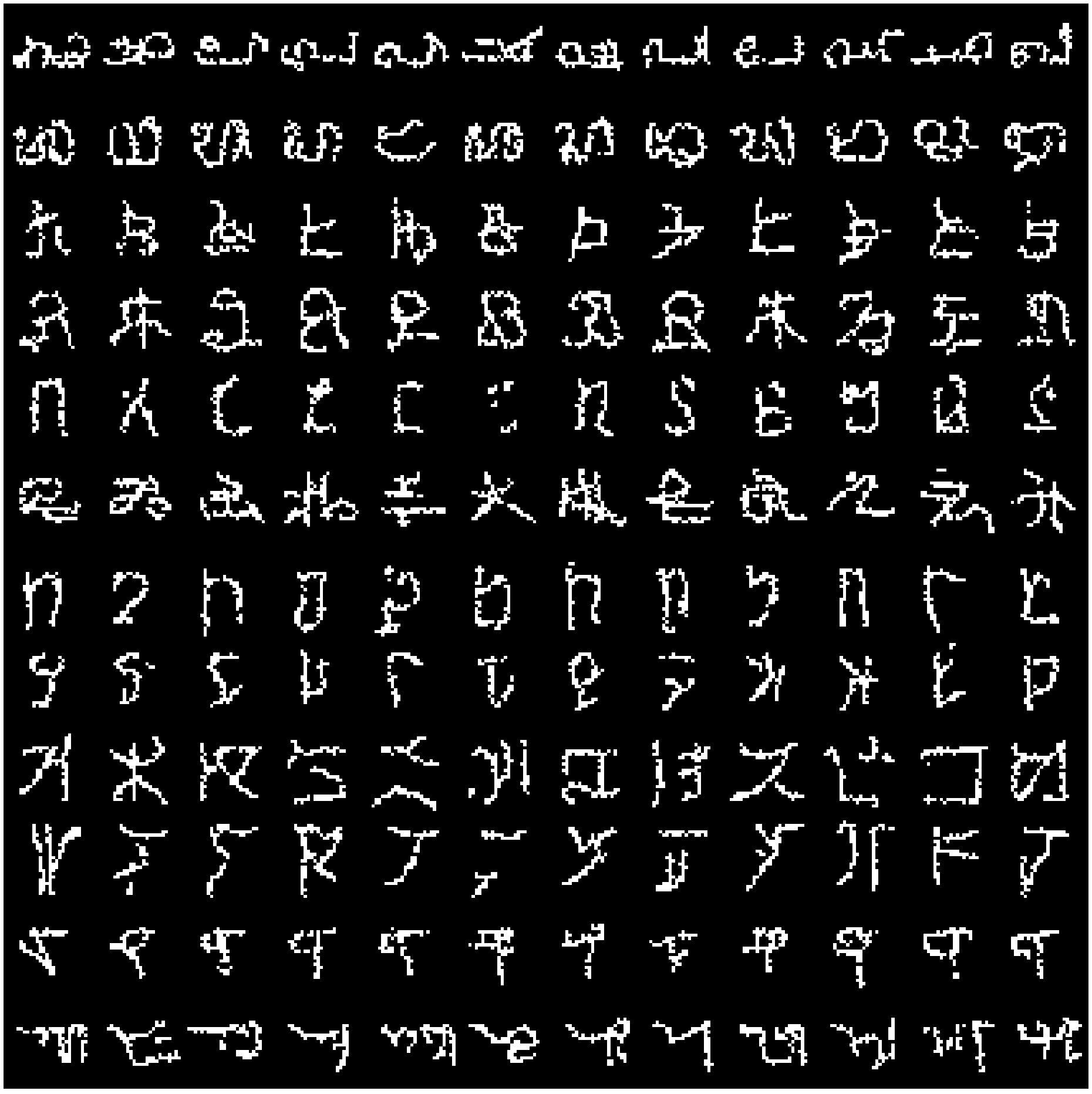}
    \includegraphics[width=6cm]{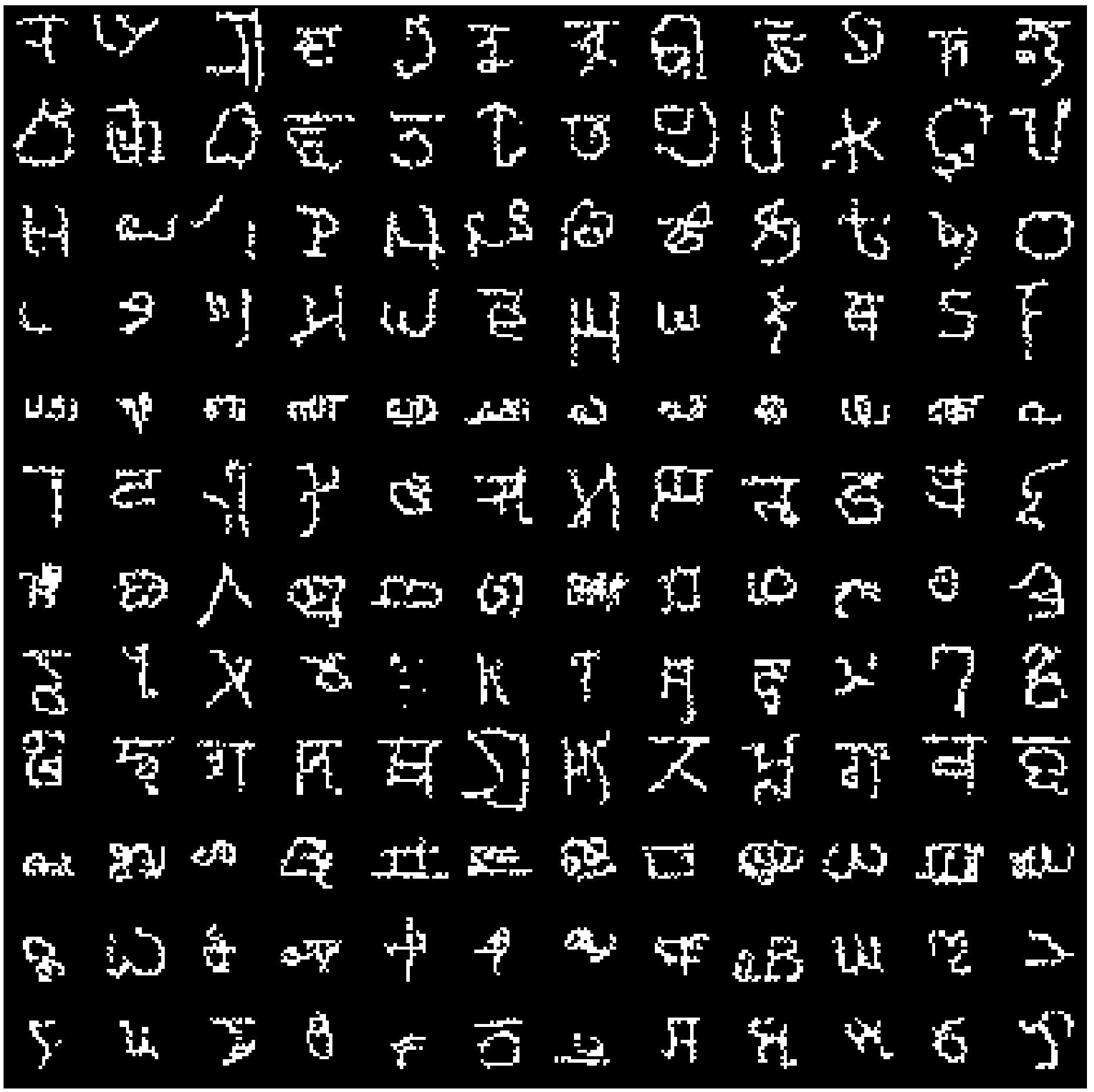}
    \caption{Generated samples for a PixelVAE++ trained on Omniglot. Each row is generated from latent variables sampled from the prior distribution. The prior distributions are RBM (left) and Gaussian (right).}
    \label{fig:omni_gen}
\end{figure}

For MNIST and Omniglot, the generated samples are fairly sharp (Fig.~\ref{fig:mnist_gen} and Fig.~\ref{fig:omni_gen}). As show in figure \ref{fig:mnist_gen}, samples from the model with RBM prior are generally fall in the same mode but occasionally change, however, the same change occurs more frequently when the prior is Gaussian.

\begin{figure}[htbp]
    \centering
    \includegraphics[width=6cm]{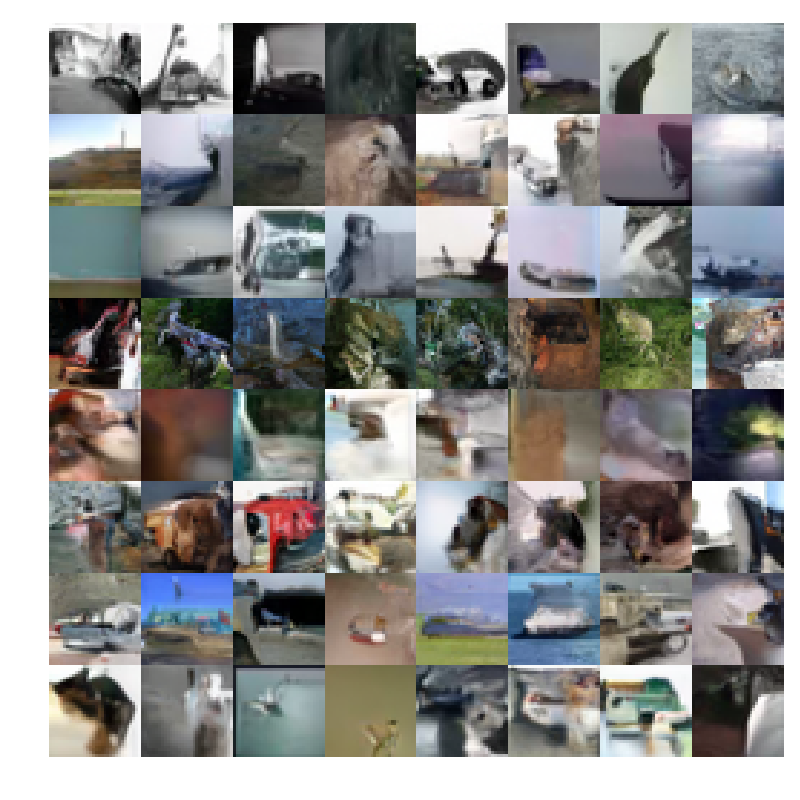}
    \includegraphics[width=6cm]{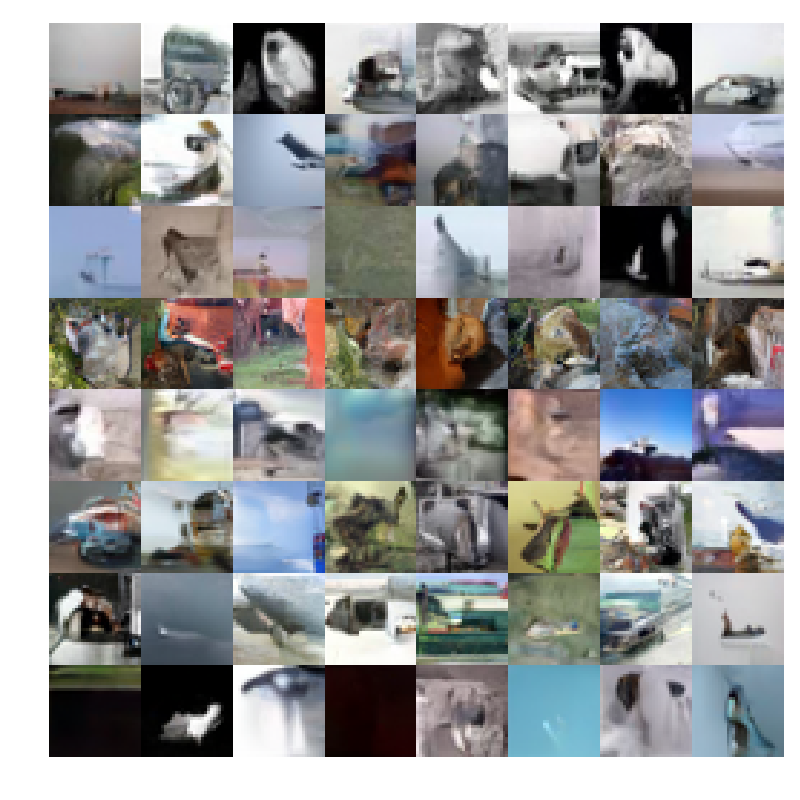}
    \caption{Generated samples. Each row corresponds to the same latent variable. Out of 128 row, we picked 8 with the smallest (left) and largest (right) mutual energy distance \cite{salimans2017pixelcnn++}.}
    \label{fig:gencifar}
\end{figure}

For CIFAR-10, there is hardly any similarity between the images. To visualize the similarity of the images (if any), we generated 128 RBM samples and 8 images per sample in each row and then sorted the rows based on the mutual energy distance \cite{salimans2017pixelcnn++}. We put the first 8 rows with the smallest mutual energy distance in the left panel and the last 8 rows on the right panel of figure \ref{fig:gencifar}. We also experimented with other metrics such as mutual $L_1$ and $L_2$ distances. In all cases there is hardly any similarity between the generated samples as in figure \ref{fig:gencifar}. While discrete latent variables capture the structure of the data well in smaller data sets (MNIST, Omniglot), neither the discrete nor continuous variables capture the structure in the CIFAR-10 data set.

\end{document}